\documentclass[10pt,twocolumn,letterpaper]{article}

\usepackage{cvpr}
\usepackage{times}
\usepackage{epsfig}
\usepackage{graphicx}
\usepackage{amsmath}
\usepackage{amssymb}
\usepackage[dvipsnames]{xcolor}
\usepackage{newtxmath}
\DeclareMathAlphabet{\mathpzc}{T1}{pzc}{m}{it}

\usepackage[pagebackref=true,breaklinks=true,colorlinks,bookmarks=false]{hyperref}

\cvprfinalcopy 

\def\cvprPaperID{830} 

\ifcvprfinal\fi
\begin{document}

\title{GNN3DMOT: Graph Neural Network for 3D Multi-Object Tracking \\ with Multi-Feature Learning}

\author{Xinshuo Weng, Yongxin Wang, Yunze Man, Kris Kitani\\
Robotics Institute \\
Carnegie Mellon University \\
{\tt\small \{xinshuow, yongxinw, yman, kkitani\}@cs.cmu.edu}
}

\maketitle
\thispagestyle{empty}

\begin{abstract}
3D Multi-object tracking (MOT) is crucial to autonomous systems. Recent work uses a standard tracking-by-detection pipeline, where feature extraction is first performed independently for each object in order to compute an affinity matrix. Then the affinity matrix is passed to the Hungarian algorithm for data association. A key process of this standard pipeline is to learn discriminative features for different objects in order to reduce confusion during data association. In this work, we propose two techniques to improve the discriminative feature learning for MOT: (1) instead of obtaining features for each object independently, we propose a novel feature interaction mechanism by introducing the Graph Neural Network. As a result, the feature of one object is informed of the features of other objects so that the object feature can lean towards the object with similar feature (i.e., object probably with a same ID) and deviate from objects with dissimilar features (i.e., object probably with different IDs), leading to a more discriminative feature for each object; (2) instead of obtaining the feature from either 2D or 3D space in prior work, we propose a novel joint feature extractor to learn appearance and motion features from 2D and 3D space simultaneously. As features from different modalities often have complementary information, the joint feature can be more discriminate than feature from each individual modality. To ensure that the joint feature extractor does not heavily rely on one modality, we also propose an ensemble training paradigm. Through extensive evaluation, our proposed method achieves state-of-the-art performance on KITTI and nuScenes 3D MOT benchmarks. Our code will be made available at \url{https://github.com/xinshuoweng/GNN3DMOT}
\end{abstract}

\vspace{-0.25cm}
\section{Introduction}
\vspace{-0.15cm}

Multi-object tracking (MOT) is an indispensable component of many applications such as autonomous driving \cite{Luo2018, Wang2018, Weng2020, Weng20202} and robot collision prediction \cite{Manglik2019}. Recent work approaches MOT in an online manner with a tracking-by-detection \cite{Bewley2016, Weng2019_3dmot} pipeline, where an object detector \cite{Shi2019, Weng2019, Ren2015, Lee2016, Weng20182} is applied to all frames and feature is extracted \emph{independently} from each detected object. Then the pairwise feature similarity is computed between objects and used to solve the MOT with a Hungarian algorithm \cite{WKuhn1955}. The key process of this pipeline is to learn discriminative features for objects with different identities.

\begin{figure}[t]
\begin{center}
\includegraphics[trim=0cm 1.5cm 9.3cm 0cm, clip=true, width=\linewidth]{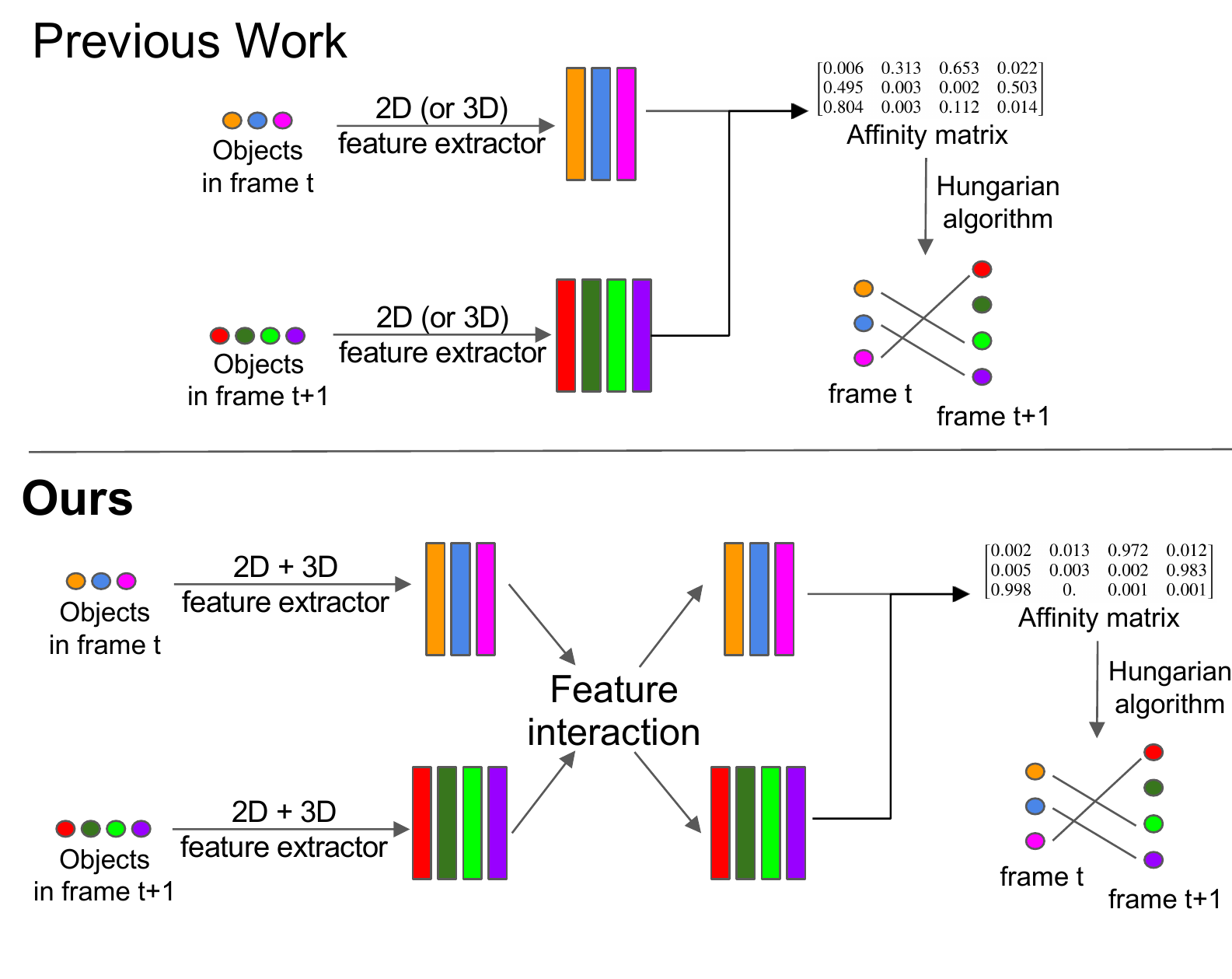}
\end{center}
  \vspace{-0.9cm}
  \caption{(\textbf{Top}): Prior work often employs a 2D or 3D feature extractor and obtain the feature independently from each object. (\textbf{Bottom}): Our work proposes a joint 2D and 3D feature extractor and a feature interaction mechanism to improve the discriminative feature learning for data association in MOT.}
  \label{fig:teaser}
  \vspace{-0.3cm}
\end{figure}
    
Our observation is that the feature extraction in prior work is always independent for each object as shown in Figure \ref{fig:teaser} (Top) and there is no interaction. For example, an object's 2D appearance feature is computed only from its own image patch, not involving with features of other objects. We found that \emph{this independent feature extraction is sub-optimal for discriminative feature learning}. This is reasonable as the feature similarity of different objects should be dependent in MOT, given the fact that an object in current frame can be matched to at most one object in previous frame. In other words, if the pairwise feature similarity of two objects is increased, then the pairwise feature similarity of any one of these two objects with all other different objects should be decreased to avoid confusion for matching. 

Based on the observation, we propose a novel \emph{feature interaction mechanism} for MOT as shown in Figure \ref{fig:teaser} (Bottom). We achieve this by introducing the Graph Neural Networks (GNNs). To the best of our knowledge, our work is the first applying the GNNs to MOT. Specifically, we construct a graph with each node being the object feature. Then, at every layer of the GNNs, each node can update its feature by aggregating features from other nodes. This node feature aggregation process is useful because each object feature is now not isolated and can be adapted with respect to other object features. We observe that, after a few GNN layers, the computed affinity matrix becomes more and more discriminative than the affinity matrix obtained without feature interaction.

In addition to the feature interaction, another primary question for discriminative feature learning in MOT is about feature selection, \emph{i.e.}, ``what type of feature should we learn?". Among different features, motion and appearance are proved to be the most useful features. Although prior works \cite{Wojke2017, Li2019, Zhang2019, Baser2019} have explored using both appearance and motion features, they only focus on either 2D or 3D space as shown in Figure \ref{fig:teaser} (top). That means, prior works use only 2D feature when approaching the 2D MOT or use only 3D feature when approaching the 3D MOT. However, this is not optimal as we know that 2D and 3D information are complementary. For example, two objects can be very close in the image but actually at a distance in 3D space because of depth discrepancy. As a result, the 3D motion feature is more discriminative in this case. On the other hand, 3D detection might not be very accurate for objects at a large distance to the camera and thus 3D motion can be very noisy. In this case, the 2D motion feature might be more discriminative.

To this end, we also propose a novel feature extractor that jointly learns motion and appearance features from both 2D and 3D space as shown in Figure \ref{fig:teaser} (bottom). Specifically, the joint feature extractor has four branches with each branch being responsible for 2D appearance, 2D motion, 3D appearance and 3D motion feature, respectively. Features from all four branches are fused before feeding into the GNNs for feature interaction. To ensure that the network does not heavily rely on one branch, we follow the concept of Dropout \cite{Srivastava2014} and propose an ensemble training paradigm, allowing the network randomly turning off branches during training. As a result, our network can learn discriminative features on all branches. 

\begin{figure*}
\begin{center}

\includegraphics[trim=0cm 0cm 0.2cm 0.1cm, clip=true, width=\linewidth]{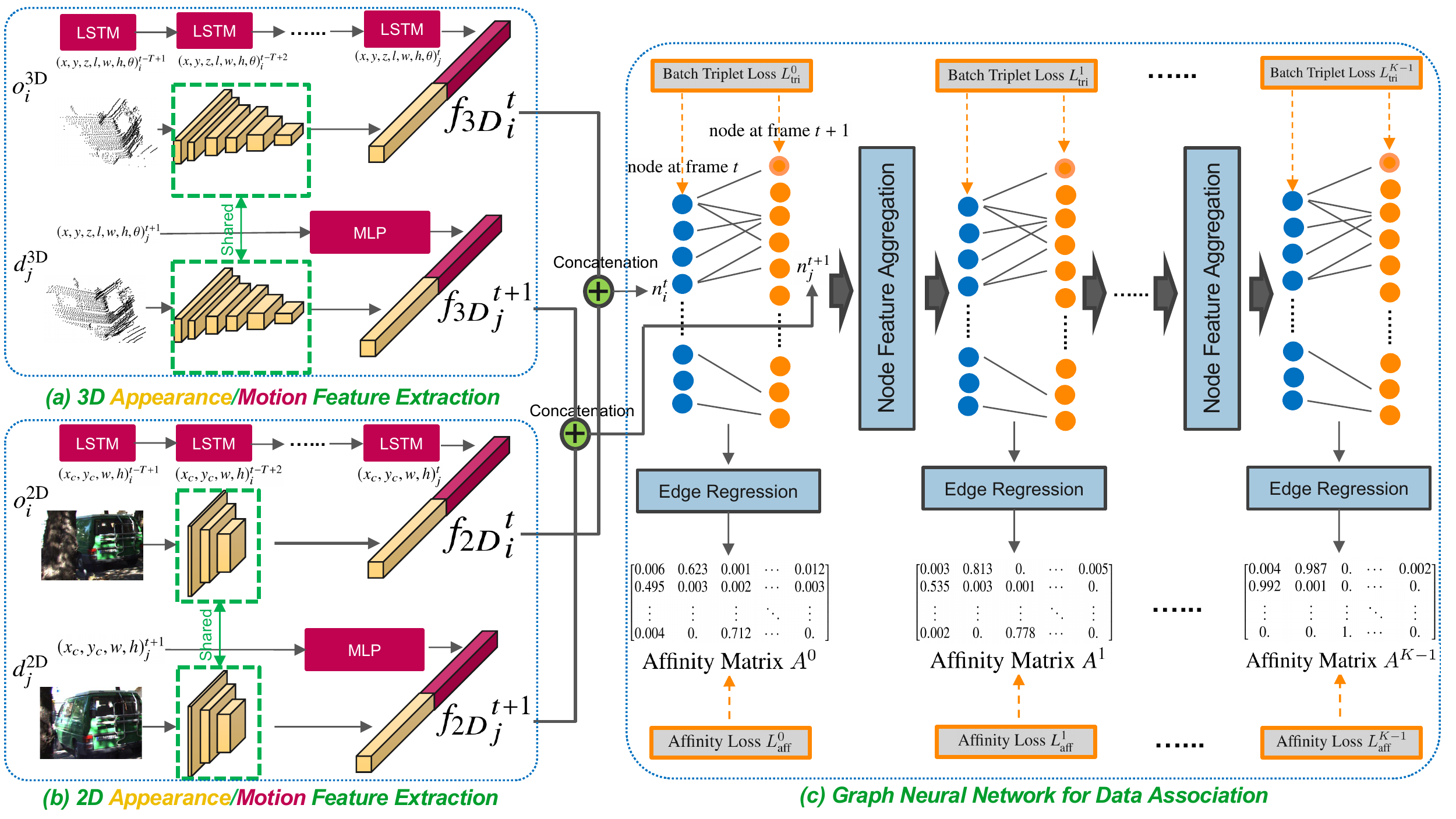}
\end{center}
\vspace{-0.5cm}
\caption{\textbf{Proposed Network.} (a)(b) Our proposed joint feature extractor obtains the feature for tracked objects $o_i$ in frame $t$ and detected objects $d_j$ in frame $t$+$1$ by utilizing the appearance and motion information from both 2D and 3D space; (c) We fuse the object features from different branches and construct a graph with the node being the object feature. Then, in every layer of the GNN, the node features are iteratively updated with the node feature aggregation technique and the affinity matrix is computed via the edge regression module. To train the entire network, we employ batch triplet loss on the node feature and affinity loss on the predicted affinity matrix in all layers.} 
\label{fig:pipeline}
\vspace{-0.2cm}
\end{figure*}

Our entire network shown in Figure \ref{fig:pipeline} is end-to-end trainable. We summarize our contributions as follows: (1) We propose a novel feature interaction mechanism for MOT by introducing the GNNs; (2) We propose a novel feature extractor along with an ensemble training paradigm to learns discriminative motion and appearance features from both 2D and 3D; (3) We achieve state-of-the-art performance on two standard 3D MOT benchmarks and also a competitive performance on the corresponding 2D MOT benchmarks.


\section{Related Work}

\noindent\textbf{Online Multi-Object Tracking.} Recent work approaches online MOT using a tracking-by-detection pipeline, where the performance is mostly affected by two factors: object detection quality and discriminative feature learning. After the affinity matrix is computed based on the pairwise similarity of learned discriminative feature, online MOT can be solved as a bipartite matching problem using the Hungarian algorithm \cite{WKuhn1955}. For a fair comparison with others, prior work often uses the same detection results so that the factor of the object detection quality can be eliminated.

To obtain discriminative feature, prior work mostly focuses on the feature selection. Among different features, it turns out that motion and appearance are the most discriminative features. Early work employs hand-crafted features such as spatial distance \cite{Pirsiavash2015} and Intersection of Union (IoU) \cite{Bochinski2017, Li2019} as the motion feature, and use color histograms \cite{Zhang2008} as the appearance feature. Recent works \cite{Sun2017, Baser2019, Zhang2019, Frossard2018, Wojke2017, yujhe2020} often employ the Convolutional Neural Networks (CNNs) to extract the appearance feature. For the motion feature, many filter-based methods \cite{Weng2019_3dmot, Bewley2016} and deep learning based methods \cite{Zhang2019, Baser2019} have been proposed. Although prior works \cite{Wojke2017, Li2019, Zhang2019, Baser2019} have explored using both motion and appearance features, they have been only focusing on either 2D or 3D space, which might lead to failure of tracking if the feature from 2D or 3D is not robust at certain frames. In contrast to prior work, we propose a novel feature extractor with four branches that jointly learns motion and appearance features from both 2D and 3D space. As a result, our method can compensate for the inaccuracy of the feature in one branch with features from other branches.

Perhaps \cite{Zhang2019_robust} is the closest to our work in terms of the feature selection as \cite{Zhang2019_robust} also proposes to jointly learn the 2D and 3D features. However, our work differs from \cite{Zhang2019_robust} as follows: (1) \cite{Zhang2019_robust} only uses the appearance feature without leveraging any motion cue. We observe that, when using both motion and appearance features, performance can be improved significantly; (2) With our proposed ensemble training paradigm, the network can be enhanced to extract high-quality features for all four branches. However, \cite{Zhang2019_robust} simply learns 2D and 3D appearance features simultaneously, which might lead to one feature dominating the other, which violates the purpose of multi-feature learning; (3) The last but most important is that our work also proposes a feature interaction mechanism for discriminative feature learning by introducing the GNNs while \cite{Zhang2019_robust} does not.

\vspace{1.5mm}\noindent\textbf{Graph Neural Networks.} In addition to the feature selection, we also propose a novel feature interaction mechanism for discriminative feature learning in MOT, which is achieved by introducing the GNNs. GNNs was first proposed by \cite{Gori2005} to directly process graph-structured data using neural networks. The major component of the GNNs is the node feature aggregation technique, with which node can update its feature by interacting with other nodes. With this technique, significant success has been achieved in many fields using GNNs such as semantic segmentation \cite{Chen2019, Zhang2019_graph}, action recognition \cite{Li2019_action, Cheng2019, Zhao2019, Wang20182}, single object tracking \cite{Gao2019}, person re-identification \cite{Wu2019}, point cloud classification and segmentation \cite{Wang2018_edgeconv}. 

Although GNNs have shown promising performance in many fields, there is no existing work that applies GNNs to MOT. To the best of our knowledge, our work is the first attempt using GNNs for online MOT. With the node aggregation technique of the GNNs, our proposed method can iteratively evolve the object features so that the feature of different objects can more discriminative. Our work is significantly different from prior work in which object features are isolated and independent of other objects. Perhaps the relation network proposed in \cite{Hu2018} is the closest to our work in terms of modeling the feature interaction. However, the feature interaction in \cite{Hu2018} only exists in the spatial domain to encode context information for object detection. Although a temporal relation network is proposed in the follow-up work \cite{Xu2019_relation}, the feature of a tracked object is only aggregating from its past trajectory and no interaction with other object features exist. In contrast, our work proposes a generic feature interaction framework that can model any kind of interaction in both spatial and temporal domains and is applicable for features from different modalities.


\section{Approach}

The goal of online MOT is to associate existing tracked objects from previous frame with new detected objects in current frame. Given $M$ tracked objects $o_i \in O$ at frame $t$ where $i\in \{1, 2, \cdots, M\}$ and also $N$ detected objects $d_j \in D$ in frame $t$+$1$ where $j\in \{1, 2, \cdots, N\}$, we want to learn discriminative feature from $O$ and $D$ and then find the correct matching based on the pairwise feature similarity. 

In Figure \ref{fig:pipeline}, our entire network consists of: (a) a 3D appearance and motion feature extractor; (b) a 2D appearance and motion feature extractor. Both 2D and 3D feature extractors are applied to all objects in $O$ and $D$ and then the extracted features are fused together, (c) a graph neural network that takes the fused object feature as input and constructs a graph with node being the object feature in frame $t$ and $t$+$1$. Then, the graph neural network iteratively aggregates the node feature from the neighborhood and computes the affinity matrix for matching using edge regression.

To apply the online MOT to an entire video at inference time, an object detector must be applied to all frames in advance. As our 2D and 3D feature extractors need object detection correspondences in 2D and 3D space, it is nontrivial to obtain the 2D detections and 3D detections separately and then obtain the detection correspondences. Instead, we only use a 3D object detector to obtain 3D detections and then 2D detections are projected from the 3D detections given the camera projection matrix. Following \cite{Shi2019, Weng2019}, we parameterize the 3D detection as a tuple of $d^{3D}$=$\{x, y, z, l, w, h, \theta\}$ where ($x, y, z$) denotes the object center in 3D space, $(l, w, h)$ denotes the object size and $\theta$ is the heading angle. For 2D detection, we parameterize it as a tuple of $d^{2D}$=$\{x_c, y_c, w, h\}$ where ($x_c, y_c$) is object center in 2D space and ($w$, $h$) denotes width and height. For tracked objects $O$, we use the same parameterization except for having an additional assigned ID $I$, \emph{i.e.}, $o^{3D}$=$\{x, y, z, l, w, h, \theta, I\}$ and $o^{2D}$=$\{x_c, y_c, w, h, I\}$.

\subsection{Joint 2D and 3D Feature Extractor}

To utilize the information for different modalities and learn discriminative feature, our proposed joint feature extractor with four branches leverages appearance and motion features from both 2D and 3D space, where two branches perform the 3D appearance and motion feature extraction and other two branches perform the 2D feature extraction. 

\vspace{1.5mm}\noindent\textbf{3D Appearance/Motion Feature Extraction.} As shown in Figure \ref{fig:pipeline} (a), given a detected object $d_j^{\mathrm{3D}}$ in frame $t$+$1$ or a tracked object $o_i^{\mathrm{3D}}$ in frame $t$, we want to obtain the corresponding 3D feature ${f_{\mathrm{3D}}}_i^t$ and ${f_{\mathrm{3D}}}_j^{t+1}$ including both appearance and motion information. For appearance branch, we use the LiDAR point cloud as the appearance cue. We first extract the point cloud enclosed by the 3D detection box and then apply the PointNet \cite{Cherabier2017, Qi2017} to obtain the feature. For motion branch, we directly use the 3D detection box as the motion cue. Note that we use different 3D motion feature extractor for tracked and detected objects as tracked objects have associated trajectory in past frames while detected objects do not have. For tracked object $o_i^{\mathrm{3D}}$, we apply the LSTMs that take into the object's 3D detections in past $T$ frames to obtain the feature. For detected object $d_j^{\mathrm{3D}}$, we use a 2-layer MLP (Multi-Layer Perceptron) that takes the detection in frame $t$+$1$ as input to extract the feature. The final 3D feature ${f_{\mathrm{3D}}}_i^t$ and ${f_{\mathrm{3D}}}_j^{t+1}$ for tracked and detected objects is obtained by concatenating the 3D motion and appearance features. To balance the contribution of the motion and appearance features, we force the final motion and appearance feature vectors to have the same dimensionality.

\vspace{1.5mm}\noindent\textbf{2D Appearance/Motion Feature Extraction.} As in Figure \ref{fig:pipeline} (b), the structure of 2D feature extractor is very similar to the 3D feature extractor explained above except for two aspects: (1) objects $o_i^{\mathrm{2D}}$ or $d_j^{\mathrm{2D}}$ are parameterized as 2D box ($x_c, y_c, w, h$) in instead of the 3D box. Therefore, the input to motion branch is different; (2) for appearance branch, we use image patch as the appearance cue, which is cropped from the entire image based on the 2D detections. To process the image patch and obtain the 2D appearance feature, we use CNNs (\emph{e.g.}, VGGNet~\cite{Simonyan2014} or ResNet~\cite{He2016}). The final 2D feature ${f_{\mathrm{2D}}}_i^t$ and ${f_{\mathrm{2D}}}_j^{t+1}$ is obtained by concatenating the 2D motion and appearance features.

\vspace{1.5mm}\noindent\textbf{Feature Fusion.} Before feeding the object feature into the graph neural network, we need to fuse the feature obtained from the 2D and 3D feature extractors. We have tried two different fusion operators: (1) concatenate the 2D and 3D features; (2) add the 2D and 3D features together. Using the ``add" fusion operator is feasible because we also force the 2D and 3D features (\emph{e.g.}, ${f_{\mathrm{2D}}}_i^t$ and ${f_{\mathrm{3D}}}_i^t$) to have the same dimensionality. We will show how different fusion operator affects the performance in the experiments. We use the concatenation as the fusion operator in our final network.

\vspace{1.5mm}\noindent\textbf{Ensemble Training Paradigm.} As our network has four branches of feature extractor and one branch may dominate the others, which violates the purpose of multi-feature learning. To avoid such cases, we propose an ensemble training paradigm. Similar to the concept of the Dropout \cite{Srivastava2014}, we randomly drop one to three branches (\emph{i.e.}, keep at least one) during every iteration of the training. Specifically, we create two random generators. The first random generator produces 0 (``not drop") or 1 (``drop") with a ratio $r$ of producing ``drop", where $r$ is a scalar between 0 and 1. In the case of ``drop", the second random generator produces a random integer between 1 to 14, which controls which combination of branches should be dropped. For example, the dropped branches can be a combination of 2D motion and 3D appearance branches.

\subsection{Graph Neural Network for Data Association}

\noindent\textbf{Graph Construction.} After feature fusion, we should have $M$ features for tracked objects in frame $t$ and also $N$ features for detected objects in frame $t$+$1$. We then construct a graph with each node being the object feature. In total, we have $M$+$N$ nodes in the graph as shown in Figure \ref{fig:pipeline} (c). We then define the neighborhood of the node (\emph{i.e.}, edges in the graph). One simple way is to have an edge between each pair of nodes, which results in a fully-connected graph and can be computationally expensive. Instead of using this simple edge construction, we utilize prior knowledge about online MOT, where the matching should only happen across frames (\emph{i.e.}, not within the same frame). Specifically, we construct the edge only between the pair of nodes in different frames. Also, for any tracked object $o_i$ in frame $t$, the possible matched detection $d_j$ in frame $t$+$1$ is most likely located in the nearby location. Therefore, we construct the edge only if two nodes' detection centers have distance less than $\mathrm{Dist}_{\mathrm{max}}^{\mathrm{3D}}$ meters in 3D space and $\mathrm{Dist}_{\mathrm{max}}^{\mathrm{2D}}$ pixels in the image. As a result, we have a sparse edge connection across frames in our final network as shown in Figure \ref{fig:pipeline} (c).

\vspace{1.5mm}\noindent\textbf{Edge Regression.} To solve the online MOT, we need to compute the $M\times N$ affinity matrix $A$ based on the pairwise similarity of the features extracted from $M$ tracked objects in frame $t$ and $N$ detected objects in frame $t$+$1$. In the context of GNN, we call this process as edge regression. We have tried three metrics for measuring the feature similarity. The first two are cosine similarity and negative L2 distance, which are conventional metrics used in the MOT community. The third one is to employ a two-layer MLP that takes the difference of two node features as input and outputs a scalar value between 0 to 1 as the pairwise similarity score:

\vspace{-0.3cm}
\begin{equation}
    A_{ij} = \mathrm{Sigmoid} (\sigma_2 (\mathrm{ReLU} (\sigma_1 (n_i^t - n_j^{t+1})))),
    \vspace{-0.1cm}
\end{equation}

\noindent where $\sigma_1$ and $\sigma_2$ are two different linear layers. In addition, $n_i^t$ and $n_j^{t+1}$ are two node features in different frames where $i \in \{1, 2, \cdots, M\}$, $j \in \{1, 2, \cdots, N\}$, In our final network, we use the MLP as the metric for edge regression and we will show how performance is affected by different metrics in the experiments.

\vspace{1.5mm}\noindent\textbf{Node Feature Aggregation.} To model feature interaction in GNN, we iteratively update the node feature by aggregating features from the neighborhood (\emph{i.e.}, nodes connected by the edge) in every layer of the GNN as shown in Figure \ref{fig:pipeline} (c). To comprehensively analyze how different types of node aggregation rules affects the performance of the MOT, we study four rules used in modern GNNs (e.g., GraphConv \cite{Morris2019}, GATConv \cite{Velickovic2018}, EdgeConv \cite{Wang2018_edgeconv}, \emph{etc}) as below:

\vspace{-0.6cm}
\begin{align}
\begin{split}\label{eq:type1}
    (\mathrm{Type \ 1}) \ \ \ {n_i^t}^{\prime} ={}& \sideset{}{_{j \in \mathpzc{N}(i)}}\sum \sigma_3 (n_j^{t+1}),
\end{split}\\
\begin{split}\label{eq:type2}
    (\mathrm{Type \ 2}) \ \ \ {n_i^t}^{\prime} ={}& \sigma_4 (n_i^t) + \sideset{}{_{j \in \mathpzc{N}(i)}}\sum \sigma_3 (n_j^{t+1}),
\end{split}\\
\begin{split}\label{eq:type3}
    (\mathrm{Type \ 3}) \ \ \ {n_i^t}^{\prime} ={}& \sigma_4 (n_i^t) + \sideset{}{_{j \in \mathpzc{N}(i)}}\sum \sigma_3 (n_j^{t+1} - n_i^t),
\end{split}\\
\begin{split}\label{eq:type4}
    (\mathrm{Type \ 4}) \ \ \ {n_i^t}^{\prime} ={}& \sigma_4 (n_i^t) + \sideset{}{_{j \in \mathpzc{N}(i)}}\sum \sigma_3 (A_{ij} (n_j^{t+1} - n_i^t)),
\end{split}
\end{align}

\vspace{-0.2cm}
\noindent where $\mathpzc{N}(i)$ denotes a set of neighborhood nodes in frame $t$+$1$ with respect to the node $i$ in frame $t$, given the fact that edge is only defined across frames in our sparse graph construction. Also, $\sigma_3$, $\sigma_4$ are linear layers which have different weights across layers of the GNN. The weight $A_{ij}$ is obtained from the affinity matrix in the current layer. Note that before the node feature aggregation in each layer, a nonlinear ReLU operator is applied to the node features.

In type 1 rule of Eq. \ref{eq:type1}, node feature is updated by aggregating features from only the neighborhood nodes, which is limited for MOT because the feature of the node itself is forgotten after aggregation. In type 2 rule, we compensate for this limitation by adding feature of the node itself as shown in the first term of Eq. \ref{eq:type2} in addition to the features aggregation from the neighborhood. In type 3 rule of Eq. \ref{eq:type3}, feature from the neighborhood node in the second term is replaced with the difference of the features between the node itself and the neighborhood node. In type 4 rule of Eq. \ref{eq:type4}, we add an attention weight obtained from the affinity matrix to the feature aggregation in the second term so that the network can focus on the neighborhood node with a higher affinity score, \emph{i.e.}, possibly the object with the same ID. We will evaluate all four node feature aggregation rules and also the number of graph layers (\emph{i.e.}, number of times performing the node feature aggregation) in the experiments.

\subsection{Losses}

Our proposed network employs two losses in all $K$ layers during training: (1) batch triple loss $L_{\mathrm{tri}}$; (2) affinity loss $L_{\mathrm{aff}}$. We can summarize the entire loss function $L$ as below:

\vspace{-0.25cm}
\begin{equation}
    L = \sideset{}{^{K-1}_{k=0}}\sum (L_{\mathrm{tri}}^{k} + L_{\mathrm{aff}}^{k}).
    \vspace{-0.2cm}
\end{equation}

\vspace{1.5mm}\noindent\textbf{Batch Triplet Loss.} In order to learn discriminative features for matching, we first apply a batch triplet loss to node feature in every layer of the GNN. For node $n_i^t$ that has a matched node $n_j^{t+1}$ (\emph{i.e.}, the object $o_i$ has the same ID with $d_j$), the batch triplet loss in each layer is defined as:

\vspace{-0.65cm}
\begin{align}
\begin{split}\label{eq:5}
\begin{aligned}
    L_{\mathrm{tri}} &= \max (\ \ || n_i^t - n_j^{t+1} || \ - \min_{\substack{d_s \in D \\ id_i \neq id_s}} || n_i^t - n_s^{t+1} || \\
                         &- \min_{\substack{o_r \in O \\ id_r \neq id_j}} || n_r^t - n_j^{t+1} || \ + \alpha, \ 0),
\end{aligned}
\end{split}
\end{align}
\vspace{-0.5cm}

\noindent where $\alpha$ is the margin of the triplet loss. $n_s^{t+1}$ is a node in frame $t$+$1$ that has a different ID from node $n_j^{t+1}$ and $n_i^{t}$. Similarly, $n_r^{t}$ is a node in frame $t$ that has a different ID from node $n_i^{t}$ and $n_j^{t+1}$. Note that the above batch triplet loss is slightly different from the original definition as in \cite{Voigtlaender2019_mots, Alexander2019}. First, we only have one positive pair of node that has the same ID as shown in the first term $|| n_i^t - n_j^{t+1} ||$ so that there is no need to apply the max operation over a batch. For the negative pair of node, we have two symmetric terms, where the first negative term forces the node feature $n_i^t$ to be different from any node that has a different ID in frame $t$+$1$ and the second negative term forces the node feature $n_j^{t+1}$ to be different from any node that has a different ID in frame $t$. In the case that $n_i^t$ does not have a matched node in frame $t$+$1$ with the same ID, we delete the first term $|| n_i^t - n_j^{t+1} ||$ for the positive pair of node in Eq. \ref{eq:5} and only minimize the remaining two negative terms in the loss $L_{\mathrm{tri}}$. 

\vspace{2mm}\noindent\textbf{Affinity Loss.} In addition to the batch triplet loss applied to the node feature, we also employ an affinity loss $L_\mathrm{aff}$ to directly supervise the final output of the network, \emph{i.e.}, the predicted affinity matrix $A$. Our affinity loss consists of two individual losses. First, as we know that the ground truth affinity matrix $A^g$ can only have integer $0$ or $1$ on all the entries, we can formulate the prediction of the affinity matrix as a binary classification problem. Therefore, our first loss is the binary cross entropy loss $L_{\mathrm{bce}}$ that is applied on each entry of our predicted affinity matrix $A$ as shown below:

\vspace{-0.6cm}
\begin{equation}
    L_{\mathrm{bce}} = \frac{-1}{MN} \sum_i^M \sum_j^N A^g_{ij} \log A_{ij} + (1 - A^g_{ij}) \log (1 - A_{ij}).
    \vspace{-0.3cm}
\end{equation}

Also, we know that each tracked object $o_i^t$ in frame $t$ can only have either one matched detection $d_j^{t+1}$ or no match at all. In other words, each row and column of the $A^g$ can only be a one-hot vector (\emph{i.e.}, a vector with $1$ in a single entry and $0$ in all other entries) or an all-zero vector. This motivates our second loss for the affinity matrix. For all rows and columns that have a one-hot vector in $A^g$, we apply the cross entropy loss $L_{\mathrm{ce}}$ to the corresponding rows and columns of $A$. As an example shown below, the column $A^g_{\cdot j}$ in ground truth affinity matrix is a one-hot vector and the loss $L_{\mathrm{ce}}$ for the $j$th column is defined as:

\vspace{-0.3cm}
\begin{equation}
    L_{\mathrm{ce}} = \frac{-1}{M} \sum_i^M  A^g_{ij} \log ( \frac{\exp{A_{ij}}}{\sum_i^M  \exp{A_{ij}}}  ).
    \vspace{-0.1cm}
\end{equation}

We can now summarize the affinity loss $L_\mathrm{aff}$ as below:

\vspace{-0.35cm}
\begin{equation}
    L_{\mathrm{aff}} = L_{\mathrm{bce}} + L_{\mathrm{ce}}.
    \vspace{-0.1cm}
\end{equation}

\subsection{Tracking Management} 

Although the discriminative feature learning can help resolve confusion for matching, it is still possible that a tracked object is matched to a false positive detection. Also, there might be the case where a tracked object still exists but cannot find a match due to missing detection (\emph{i.e.}, false negative). To avoid such cases, a tracking management module that controls the birth and death of the objects is necessary in MOT to reduce the false positives and false negatives. We follow \cite{Bewley2016, Weng2019_3dmot} and maintain a death count and a birth count for each object. If a new object is able to find the match in $\mathrm{Bir}_{\mathrm{min}}$ frames continuously, we will then assign an ID to this object and add it to the set of tracked objects $O$. However, if this object stops finding the match before being assigned an ID, we will reset the birth count to zero. On the other hand, if a tracked object cannot find the matched detection in $\mathrm{Age}_{\mathrm{max}}$ frames, we believe that this object has disappeared and will delete it from the set of tracked objects $O$. However, if this tracked object can still find a match before being deleted, we believe that the object still exists and will reset the death count to zero. In the first frame of the video, we initialize the tracked objects $O$ as an empty set.


\begin{table*}[t]
\caption{Quantitative comparison on KITTI-Car val set. The evaluation is conducted in \textbf{3D} space using \cite{Weng2019_3dmot} 3D MOT evaluation tool.}
\vspace{0.05cm}
\centering
\resizebox{\textwidth}{!}{
\begin{tabular}{|c|c|c|c|c|c|c|c|c|}
\hline
Method       & Input Data & \textbf{sAMOTA} (\%) $\uparrow$ & AMOTA (\%) $\uparrow$ & AMOTP (\%) $\uparrow$ & MOTA (\%) $\uparrow$ & MOTP (\%) $\uparrow$ & IDS $\downarrow$ & FRAG $\downarrow$ \\
\hline
mmMOT~\cite{Zhang2019_robust} (ICCV$'$19) & 2D + 3D & 70.61 & 33.08 & 72.45 & 74.07 & 78.16 & 10 & 125\\ 
FANTrack~\cite{Baser2019} (IV$'$19) & 2D + 3D & 82.97 & 40.03 & 75.01 & 74.30 & 75.24 & 35 & 202 \\
AB3DMOT\cite{Weng2019_3dmot} (arXiv$'$19) & 3D & 91.78 & 44.26 & 77.41 & 83.35 & 78.43 & \textbf{0} & 15  \\
\hline
\textbf{Ours} & 2D + 3D & \textbf{93.68} & \textbf{45.27} & \textbf{78.10} & \textbf{84.70} & \textbf{79.03} & \textbf{0} &  \textbf{10} \\
\hline
\end{tabular}}
\vspace{-0.35cm}
\label{tab:kitti_3d}
\end{table*}

\begin{table*}[t]
\caption{Quantitative comparison on KITTI-Car test set. The evaluation is conducted in \textbf{2D} space using KITTI 2D MOT evaluation tool.}
\vspace{0.05cm}
\centering
\resizebox{\textwidth}{!}{
\begin{tabular}{|c|c|c|c|c|c|c|c|c|c|c|}
\hline
Method & Input Data & \textbf{MOTA} (\%) $\uparrow$ & MOTP (\%) $\uparrow$ & MT (\%) $\uparrow$  & ML (\%) $\downarrow$  & IDS $\downarrow$ & FRAG $\downarrow$  & FPS $\uparrow$\\
\hline
CIWT~\cite{Osep2017} (ICRA$'$17) & 2D & 75.39 & 79.25 & 49.85 & 10.31 & 165 & 660 & 2.8 \\
FANTrack~\cite{Baser2019} (IV$'$19) & 2D + 3D & 77.72     & 82.32     & 62.61     & 8.76      & 150 & 812     & 25.0 (GPU)\\
AB3DMOT\cite{Weng2019_3dmot} (arXiv$'$19) & 3D & 83.84 & 85.24 & 66.92 & 11.38 & \textbf{9} & \textbf{224} & \textbf{214.7}\\ 
BeyondPixels \cite{Sharma2018} (ICRA$'$18) & 2D & 84.24 & 85.73 & 73.23 & 2.77 & 468 & 944 & 3.33\\
3DT~\cite{Hu2019} (ICCV$'$19) & 2D & 84.52 & \textbf{85.64} & 73.38 & \textbf{2.77} & 377 & 847 & 33.3 (GPU)\\
mmMOT~\cite{Zhang2019_robust} (ICCV$'$19) & 2D + 3D & 84.77 & 85.21 & 73.23 & \textbf{2.77} & 284 & 753 & 4.8 (GPU)\\
MASS \cite{Karunasekera2019} (IEEE Access$'$19) & 2D & \textbf{85.04} & 85.53 & \textbf{74.31} & \textbf{2.77} & 301 & 744 & 100.0\\ 
\hline
\textbf{Ours} & 2D + 3D & 80.40 & 85.05 & 70.77 & 11.08 & 113 & 265 & 5.2 (GPU) \\
\textbf{Ours} + 2D detections from \cite{Ren2017} & 2D & 82.24 & 84.05 & 64.92 & 6.00 & 142 & 416 & 5.1 (GPU) \\
\hline
\end{tabular}}
\vspace{-0.6cm}
\label{tab:kitti_2d}
\end{table*}

\begin{table}[t]
\caption{Quantitative comparison on nuScenes validation set. The evaluation is conducted in \textbf{3D} space with 3D MOT evaluation tool.}
\vspace{0.05cm}
\centering
\resizebox{\hsize}{!}{
\begin{tabular}{|c|c|c|c|c|}
\hline
Method       & \textbf{sAMOTA} (\%) $\uparrow$ & AMOTA (\%) $\uparrow$ & AMOTP (\%) $\uparrow$ & MOTA (\%) $\uparrow$ \\
\hline
FANTrack~\cite{Baser2019} & 19.64 & 2.36 & 22.92 & 18.60  \\
mmMOT~\cite{Zhang2019_robust} & 23.93 & 2.11 & 21.28 & 19.82 \\ 
AB3DMOT\cite{Weng2019_3dmot} & 27.90 & 4.93 & 23.89 & 21.46 \\
\hline
\textbf{Ours} & \textbf{29.84} & \textbf{6.21} & \textbf{24.02} & \textbf{23.53} \\
\hline
\end{tabular}}
\vspace{-0.3cm}
\label{tab:nuscenes_3d}
\end{table}


\section{Experiments}
\subsection{Settings}

\noindent\textbf{Dataset.} To demonstrate the strength of our joint 2D-3D feature extractor, we evaluate our network on KITTI \cite{Geiger2012} and nuScenes \cite{Caesar2019} datasets, which provide both 2D (images and 2D boxes) and 3D data (LiDAR point cloud and 3D boxes). For KITTI, same as most prior works, we report results on the car subset for comparison. For nuScenes, we evaluate on all categories and the final performance is the mean over all categories. As the focus of this paper is 3D MOT, we report and compare 3D MOT performance on the KITTI and nuScenes datasets. Since KITTI has an official 2D MOT benchmark, we also report 2D MOT results on KITTI for reference, which is achieved by projecting our 3D MOT results to the image space.

\vspace{1.5mm}\noindent\textbf{Evaluation Metrics.} We use standard CLEAR metrics \cite{Bernardin2008} (including MOTA, MOTP, IDS, FRAG and FPS) and also the new sAMOTA, AMOTA and AMOTP metrics proposed in \cite{Weng2019_3dmot} for 3D MOT and 2D MOT evaluation. For 3D MOT evaluation, we use the evaluation tool proposed by \cite{Weng2019_3dmot}. As KITTI and nuScenes datasets do not release the ground truth of test set to users, we use the validation set for 3D MOT evaluation. For KITTI 2D MOT evaluation, we use the official KITTI 2D MOT evaluation tool \cite{Geiger2012}. In terms of the training, validation and testing split, we use the official one on nuScenes. As KITTI does not have an official split, we use the one proposed by \cite{Scheidegger2018}.

\vspace{1.5mm}\noindent\textbf{Baselines.} For 3D MOT, we compare with recent open-source 3D MOT systems such as FANTrack \cite{Baser2019}, mmMOT \cite{Zhang2019_robust} and AB3DMOT \cite{Weng2019_3dmot}, which also use the 3D LiDAR data (either directly used in 3D MOT or indirectly used in order to obtain the 3D detections for 3D MOT) for fair comparison with our 3D MOT method. 
For 2D MOT, we compare with state-of-the-art published 2D MOT systems on the KITTI MOT leaderboard.

\subsection{Implementation Details}

\noindent\textbf{3D Object Detection.} For fair comparison in KITTI, we use the same 3D detections from PointRCNN \cite{Shi2019} for all 3D MOT methods (including our method and the baselines) that require 3D detections as inputs. For 3D MOT methods that also require 2D detections, \emph{e.g.}, the 2D feature extraction branch in our method, we use the 2D projection of 3D detections from \cite{Shi2019}. For nuScenes, the same rule also applies except that the 3D detections obtained by \cite{Shi2019} is replaced with the 3D detections obtained by \cite{Zhu2019}. For data augmentation, we perturb the ground truth box during training with a ratio of $0.1$ with respect to the size of the box.

\vspace{1.5mm}\noindent\textbf{Joint Feature Extractor.} We use the feature with same dimensionality of $64$ for all four branches. For 3D appearance branch, we use the PointNet with six 1D Convolutional layers that maps the input point cloud with size of P (number of points) $\times$ 4 ($x$, $y$, $z$, reflectance) to P $\times$ 64 (4$\Rightarrow$16$\Rightarrow$32 $\Rightarrow$64$\Rightarrow$128$\Rightarrow$256$\Rightarrow$64). Then, a max pooling operation is applied along the axis of $P$ to obtain the 3D appearance feature with the dimensionality of 64. For 2D appearance branch, we resize the cropped image patch for each object to 56 $\times$ 84 and use the ResNet34 to extract the 2D appearance feature. For 2D and 3D motion branches, we use a two-layer LSTMs with a hidden size of 64 and number of past frames $T$=$5$ for tracked objects. For tracked objects which only have associated detections in past $R$ ($<$ $T$) frames, we repeat the earliest detection $T$-$R$ times so that the objects can have $T$ frames of detections. For detected objects, we employ a two-layer MLP (4$\Rightarrow$16$\Rightarrow$64 in 2D motion branch, 7$\Rightarrow$32 $\Rightarrow$64 in 3D motion branch).

\vspace{1.5mm}\noindent\textbf{Feature Fusion and Ensemble Training Paradigm.} In feature fusion, if a branch is dropped, we fill in zeros into the feature corresponding to the dropped branch before fusion so that the feature fusion module is compatible with the ensemble training paradigm. For drop ratio, we use $r=0.5$.

\vspace{1.5mm}\noindent\textbf{Graph Neural Network and Miscellaneous.} We use the $\mathrm{Dist}_{\mathrm{max}}^{\mathrm{3D}}$=$5$ and $\mathrm{Dist}_{\mathrm{max}}^{\mathrm{2D}}$=$200$ in our sparse graph construction. We use three GNN layers (\emph{i.e.}, $K$=$4$) with each layer having feature with same dimensionality. For example, when we use ``concatenate'' as the fusion operator, we will have node feature with dimensionality of $256$ in all layers of GNN. For edge regression, we use a two-layer MLP with hidden feature dimension of 256$\Rightarrow$64$\Rightarrow$1. For batch triplet loss, we use the margin $\alpha$=$10$. For the tracking management, we use $\mathrm{Age}_{\mathrm{max}}$=$4$ and $\mathrm{Bir}_{\mathrm{min}}$=$10$.

\subsection{Experimental Results} 

\noindent\textbf{Results on KITTI.} We summarize the 3D MOT and 2D MOT results on KITTI-Car dataset in Table \ref{tab:kitti_3d} and \ref{tab:kitti_2d}. For 3D MOT evaluation in Table \ref{tab:kitti_3d}, our proposed method consistently outperforms other modern 3D MOT systems in all metrics. For 2D MOT evaluation in Table \ref{tab:kitti_2d}, our network is behind prior work and only achieves $80.40$ 2D MOTA. One possible reason is that the 2D projection of 3D detection results we use has lower precision and recall than a state-of-the-art 2D detector \cite{Ren2017} used in prior work. For fair comparison, we simply replace the input 2D detections with \cite{Ren2017} while keeping all hyper-parameters fixed and show the results in the last row of Table \ref{tab:kitti_2d}. As a result, the MOTA of our proposed method is improved about 2\% without bells and whistles. We argue that it is highly possible that our proposed method can achieve higher performance on 2D MOT after hyper-parameter searching based on 2D MOT evaluation. Currently, all ablation analysis is performed on 3D MOT evaluation, meaning that the hyper-parameters of our method are only tuned for 3D MOT and not for 2D MOT.

\vspace{1.5mm}\noindent\textbf{Results on nuScenes.} In Table \ref{tab:nuscenes_3d}, our method achieves the state-of-the-art 3D MOT performance on nuScenes. As the 3D detection performance is not yet mature on nuScenes compared to KITTI, 3D MOT performance on nuScenes is consistently lower than on KITTI.

\vspace{1.5mm}\noindent\textbf{Inference Time.} Our network runs at a rate of 5.2 FPS on the KITTI test set with a single 1080Ti GPU.

\begin{figure*}
\begin{center}

\includegraphics[trim=0cm 0cm 0cm 0cm, clip=true, width=0.33\linewidth]{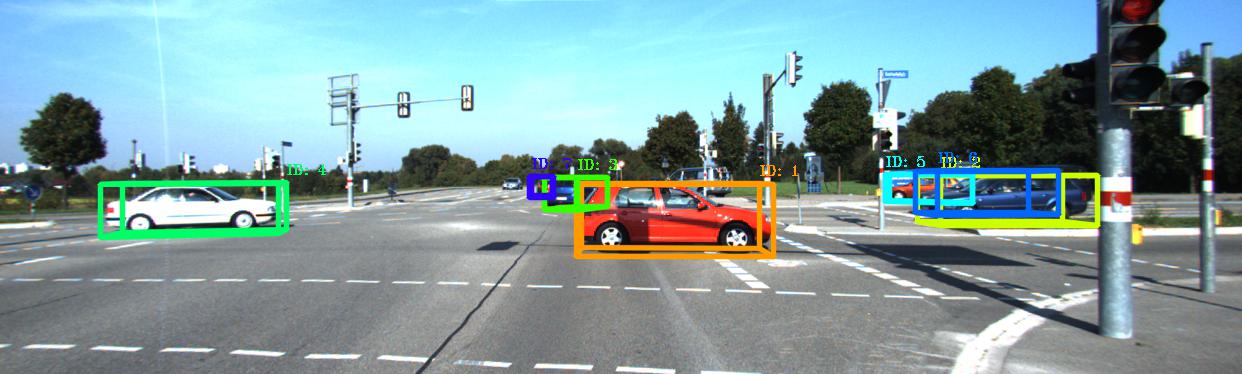}
\includegraphics[trim=0cm 0cm 0cm 0cm, clip=true, width=0.33\linewidth]{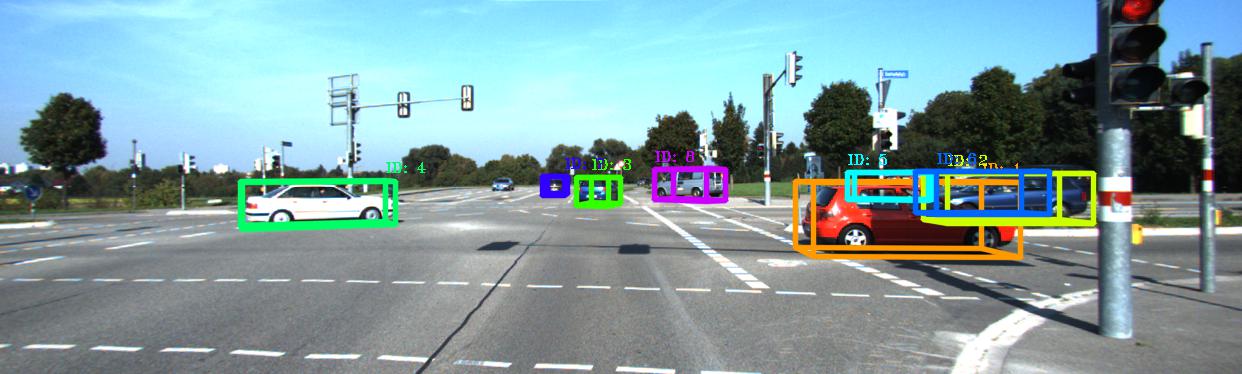}
\includegraphics[trim=0cm 0cm 0cm 0cm, clip=true, width=0.33\linewidth]{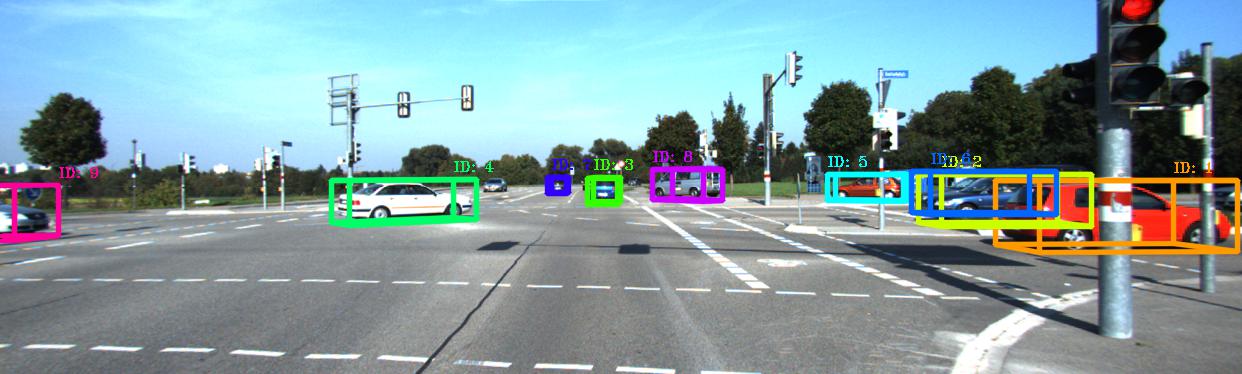}

\includegraphics[trim=0cm 0cm 0cm 0cm, clip=true, width=0.33\linewidth]{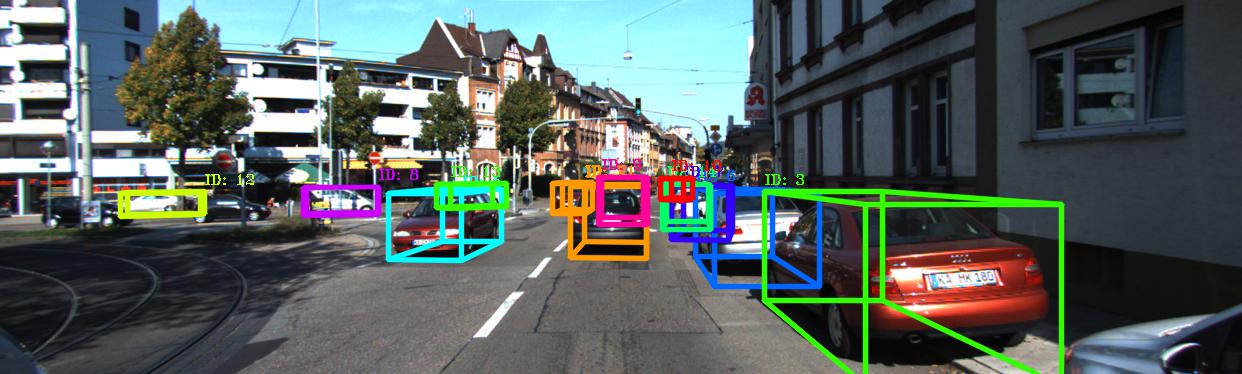}
\includegraphics[trim=0cm 0cm 0cm 0cm, clip=true, width=0.33\linewidth]{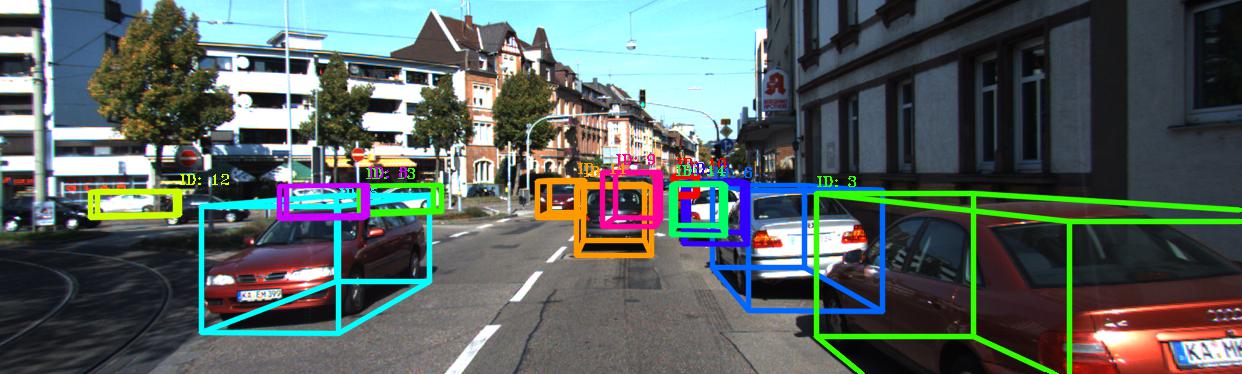}
\includegraphics[trim=0cm 0cm 0cm 0cm, clip=true, width=0.33\linewidth]{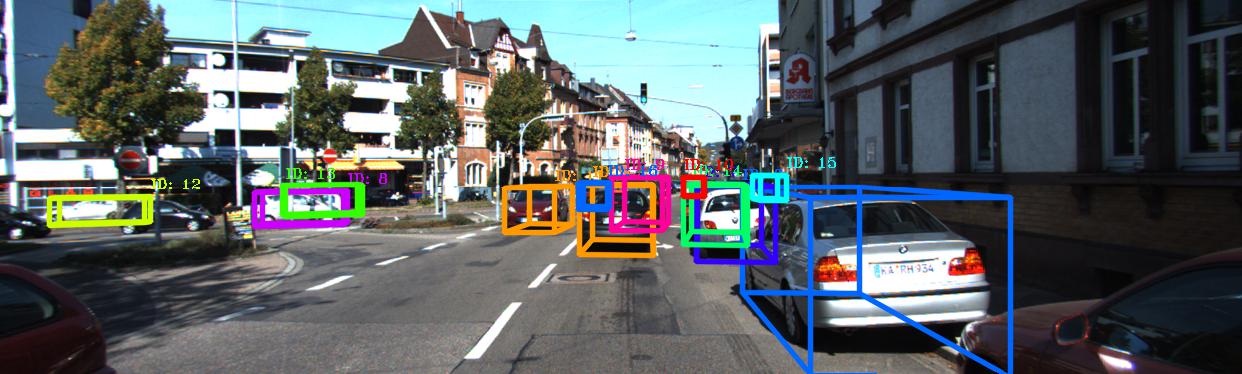}

\end{center}
\vspace{-0.45cm}
\caption{Qualitative results of our method on sequence 10 (top row) and 11 (bottom row) of the KITTI test set.}
\label{fig:qua}
\vspace{-0.3cm}
\end{figure*}

\vspace{1.5mm}\noindent\textbf{Qualitative Comparison.} We show our qualitative results on two sequences of the KITTI test set in Figure \ref{fig:qua}. 

\begin{figure*}
\begin{center}
\begin{minipage}[c]{0.33\textwidth}
    \begin{center}
    \centering
    \includegraphics[trim=0cm 0cm 0cm 0cm, clip=true, width=\linewidth]{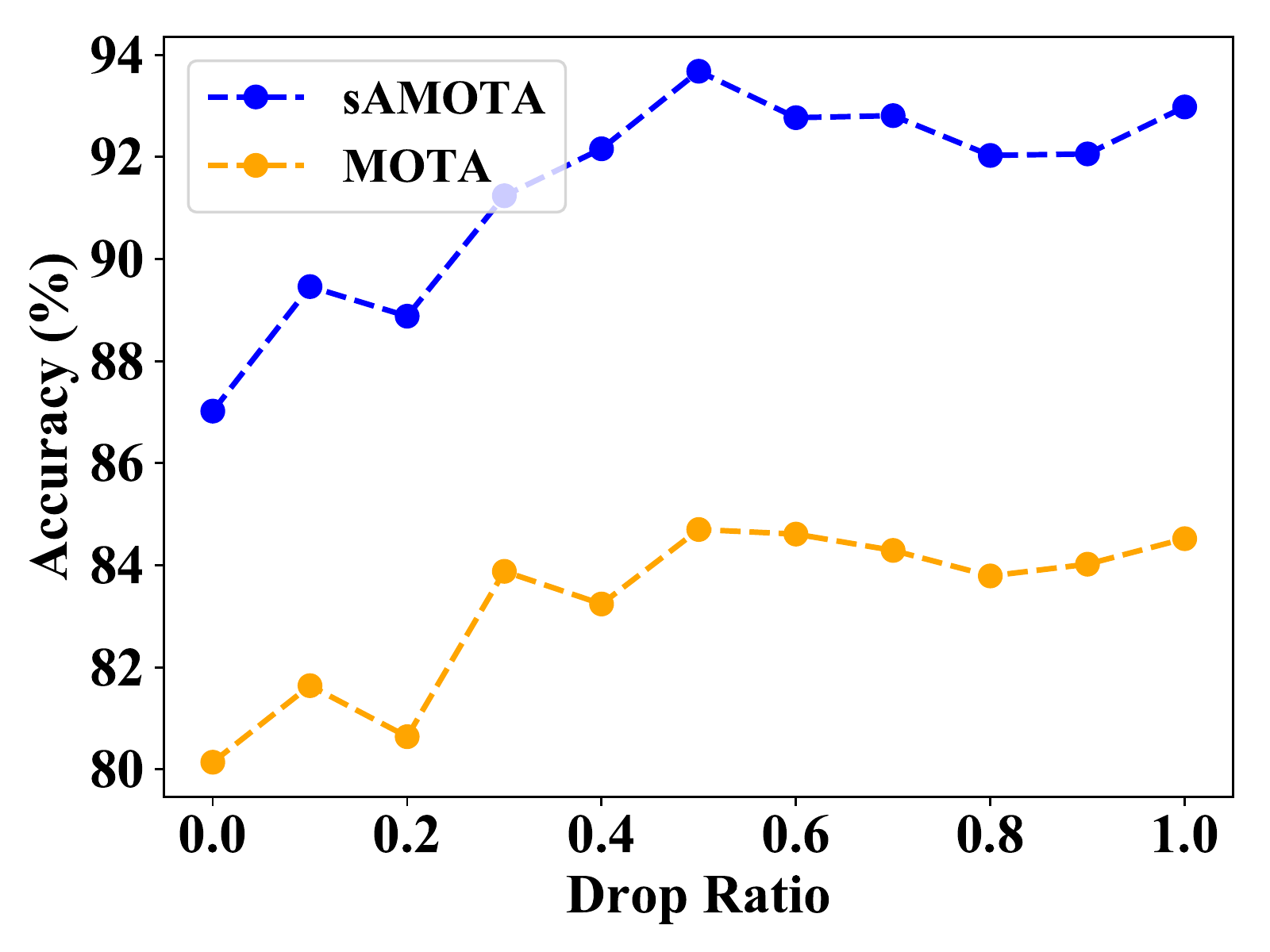}
    \vspace{-0.7cm}
    \\ (a) Accuracy v.s. Drop Ratio
    \label{fig:mota_recall} 
    \end{center}
\end{minipage}
\begin{minipage}[c]{0.33\textwidth}
    \begin{center}
    \includegraphics[trim=0cm 0cm 0cm 0cm, clip=true, width=\linewidth]{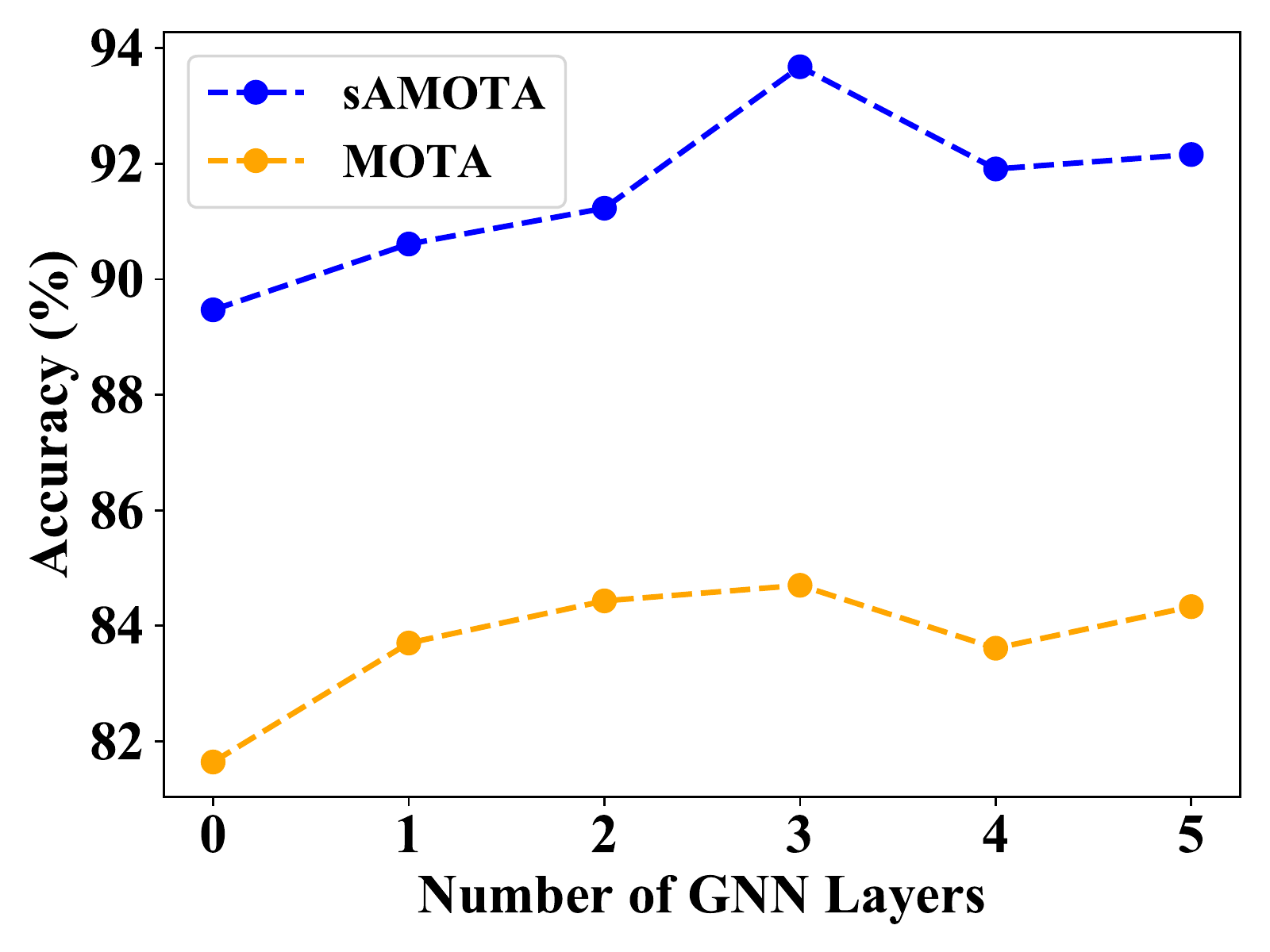}
    \vspace{-0.7cm}
    \\ (b) Accuracy v.s. Number of Layers
    \label{fig:precision_recall} 
    \end{center}
\end{minipage}
\begin{minipage}[c]{0.33\textwidth}
    \begin{center}
    \includegraphics[trim=0cm 0cm 0cm 0cm, clip=true, width=\linewidth]{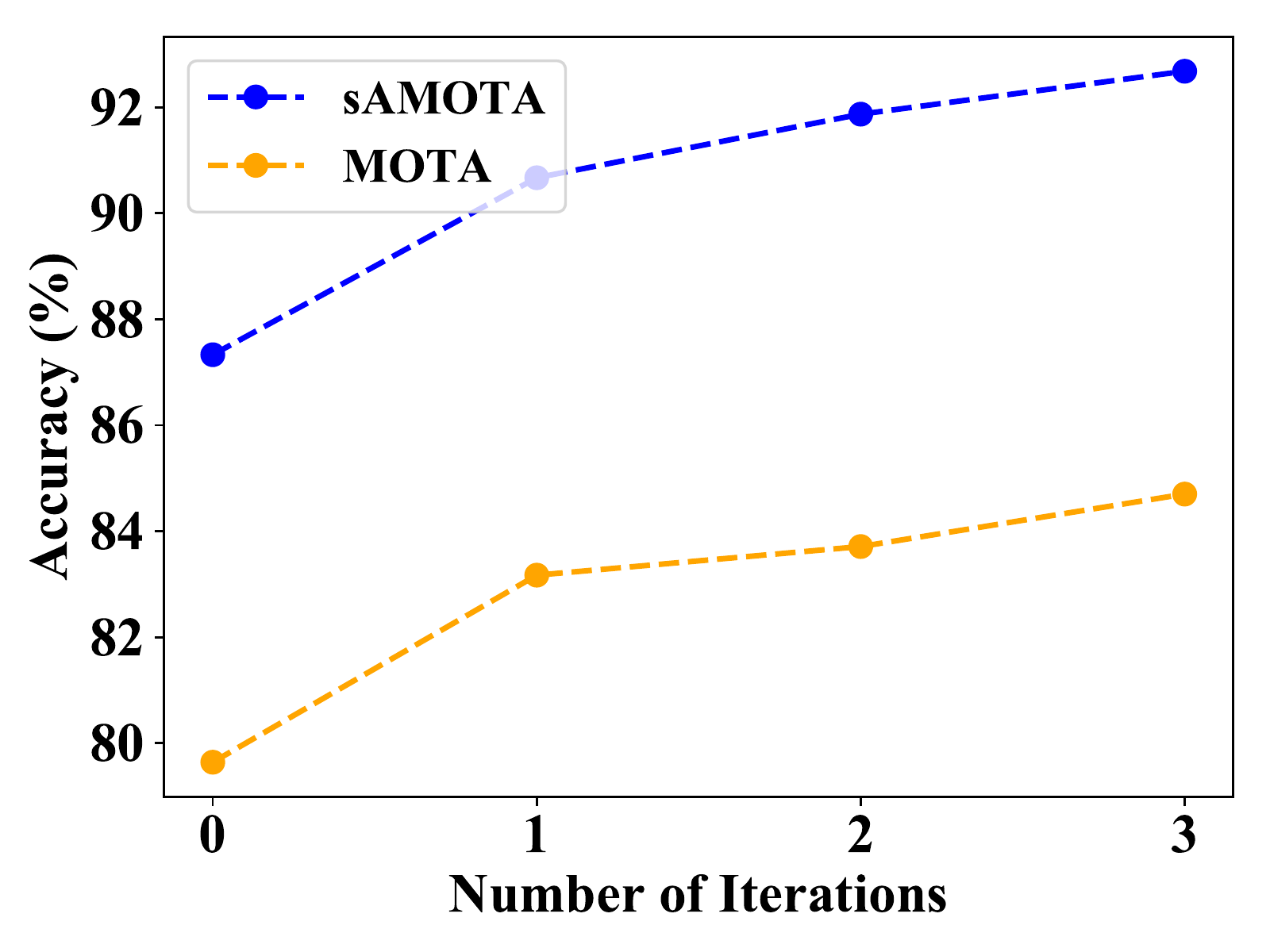}
    \vspace{-0.7cm}
    \\ (c) Accuracy vs Number of Iterations
    \label{fig:fn_recall} 
    \end{center}
\end{minipage}
\end{center}
\vspace{-0.3cm}
\caption{(a) \textbf{Effect of Ensemble Training Paradigm.} We vary the drop ratio $r$ from $0$ to $1$ with an interval of $0.1$. Results suggest that $r$=$0.5$ is the best. (b) \textbf{Effect of Number of GNN Layers.} We increase the number of layers from $0$ (\emph{i.e.}, deactivate the GNN) to $5$ and use the output from the last layer of GNN for evaluation. The highest accuracy is obtained when using three layers. (c) \textbf{Effect of Feature Interaction.} For our final network with three GNN layers, we evaluate the output of layer $0$ (\emph{i.e.}, deactivate the GNN) to layer $3$. Results suggest that the output from the last layer of the GNN achieve the highest performance.}
\label{fig:exp}
\end{figure*}

\subsection{Ablation Study}

We conduct the ablation study on KITTI-Car validation set using 3D MOT evaluation tool proposed by \cite{Weng2019_3dmot}.


\begin{table}
\caption{\textbf{Effect of Joint Feature Extractor}. Results are evaluated on KITTI-Car val set using \textbf{3D} MOT evaluation tool. Appearance and motion features are denoted as ``A'' and ``M'' respectively.}
\vspace{-0.25cm}
\begin{center}
\resizebox{\hsize}{!}{
\begin{tabular}{|c|c|c|c|c|}
\hline
Feature Extractor & \textbf{sAMOTA} (\%) $\uparrow$ & AMOTA (\%) $\uparrow$ & AMOTP (\%) $\uparrow$ & MOTA (\%) $\uparrow$ \\
\hline
2D A & 88.31 & 41.62 & 76.22 & 79.42\\
2D M & 64.24 & 23.95 & 61.13 & 54.88\\
3D A & 88.27 & 41.55 & 76.29 & 77.38 \\
3D M & 88.57 & 41.62 & 76.22 & 81.84\\
\hline
2D+3D A & 89.39 & 42.55 & 76.24 & 83.02\\
2D+3D M & 91.75 & 44.75 & 78.05 & 84.54\\
\hline
2D M+A & 90.56 & 44.39 & \textbf{78.20} & 83.15\\
3D M+A & 91.30 & 44.31 & 78.16 & 84.06\\
\hline
2D+3D M+A (\textbf{Ours}) & \textbf{93.68} & \textbf{45.27} & 78.10 & \textbf{84.70}\\
\hline
\end{tabular}}
\end{center}
\vspace{-0.65cm}
\label{tab:feature}
\end{table}

\vspace{1.5mm}\noindent\textbf{Effect of Joint Feature Extractor.} In Table \ref{tab:feature}, we evaluate the effect of each individual feature extractor and the combination of them. We show that combining features from different modalities improves performance, suggesting that different features are complementary to others.  


\begin{table}
\caption{\textbf{Effect of Feature Fusion Operators.} Results are evaluated on KITTI-Car val set using \textbf{3D} MOT evaluation tool.}
\vspace{-0.3cm}
\begin{center}
\resizebox{\hsize}{!}{
\begin{tabular}{|c|c|c|c|c|}
\hline
Fusion & \textbf{sAMOTA} (\%) $\uparrow$ & AMOTA (\%) $\uparrow$ & AMOTP (\%) $\uparrow$ & MOTA (\%) $\uparrow$ \\
\hline
Add & 89.98 & 42.97 & 75.96 & 82.55\\
Concatenate (\textbf{Ours}) & \textbf{93.68} & \textbf{45.27} & \textbf{78.10} & \textbf{84.70}\\
\hline
\end{tabular}}
\end{center}
\vspace{-0.65cm}
\label{tab:fusion}
\end{table}

\vspace{1.5mm}\noindent\textbf{Effect of Feature Fusion Operators.} In Table \ref{tab:fusion}, we show that using ``concatenate'' is better than ``add'' for fusion.


\begin{table}
\caption{\textbf{Effect of Edge Regression Modules}. Results are evaluated on KITTI-Car val set using \textbf{3D} MOT evaluation tool.}
\vspace{-0.3cm}
\begin{center}
\resizebox{\hsize}{!}{
\begin{tabular}{|c|c|c|c|c|}
\hline
Edge Regression & \textbf{sAMOTA} (\%) $\uparrow$ & AMOTA (\%) $\uparrow$ & AMOTP (\%) $\uparrow$ & MOTA (\%) $\uparrow$ \\
\hline
Negative L2 Distance & 82.26 & 41.38 & 72.42 & 70.71\\
Cosine Similarity & 87.07 & 43.18 & 72.17 & 75.46\\

MLP (\textbf{Ours}) & \textbf{93.68} & \textbf{45.27} & \textbf{78.10} & \textbf{84.70}\\
\hline
\end{tabular}}
\end{center}
\vspace{-0.65cm}
\label{tab:edge}
\end{table}

\vspace{1.5mm}\noindent\textbf{Effect of Edge Regression Modules.} In Table \ref{tab:edge}, the two-layer MLP used in our final network achieves better performance than the conventional similarity metrics. 


\begin{table}
\caption{\textbf{Effect of Node Aggregation Rules}. Results are evaluated on KITTI-Car val set using \textbf{3D} MOT evaluation tool.}
\vspace{-0.3cm}
\begin{center}
\resizebox{\hsize}{!}{
\begin{tabular}{|c|c|c|c|c|}
\hline
Node Aggregation & \textbf{sAMOTA} (\%) $\uparrow$ & AMOTA (\%) $\uparrow$ & AMOTP (\%) $\uparrow$ & MOTA (\%) $\uparrow$ \\
\hline
Type 1 & 75.61 & 32.84 & 65.81 & 67.43\\
\hline
Type 2 (SAGEConv \cite{Hamilton2017}) & 87.81 & 41.06 & 76.29 & 77.22\\
Type 2 (GCN \cite{Kipf2017}) & 89.78 & 43.37 & 78.06 & 80.67\\
Type 2 (GraphConv \cite{Morris2019_graph}) & 91.15 & 44.78 & 77.93 & 82.31\\
Type 2 (GATConv \cite{Velickovic2018_GAT}) & 91.66 & 44.57 & 77.99 & 82.37\\
Type 2 (AGNNConv \cite{Thekumparampil2018}) & 91.88 & 44.95 & 78.00 & 84.32\\
\hline
Type 3 (EdgeConv \cite{Wang2018_graph}) & 92.17 & 44.65 & 77.98 & 83.73\\
\hline
Type 4 (\textbf{Ours}) & \textbf{93.68} & \textbf{45.27} & \textbf{78.10} & \textbf{84.70}\\
\hline
\end{tabular}}
\end{center}
\vspace{-0.3cm}
\label{tab:node}
\end{table}

\vspace{1.5mm}\noindent\textbf{Effect of Node Aggregation Rules.} In Table \ref{tab:node}, we show that type 4 rule performs the best. Also, for different GNNs with type 2 rule, performance varies significantly. 


\vspace{1.5mm}\noindent\textbf{Effect of Ensemble Training Paradigm.} In Figure \ref{fig:exp} (a), we observe that using ensemble training paradigm significantly improves the performance with $r$=$0.5$ being the best.

\vspace{1.5mm}\noindent\textbf{Effect of Number of GNN Layers.} In Figure \ref{fig:exp} (b), increasing the number of GNN layers improves the performance with three GNN layers being the best. We did not experiment with GNN larger than five layers as the GNN tends to overfit when it becomes very deep.

\vspace{1.5mm}\noindent\textbf{Effect of Feature Interaction.} In Figure \ref{fig:exp} (c), we show that feature interaction in GNNs is effective as the performance increases when we use the output from a later layer.


\section{Conclusion}

We propose a 3D MOT method with a novel joint 2D-3D feature extractor and a novel feature interaction mechanism achieved by GNNs in order to improve the discriminative feature learning in MOT. Through extensive experiments, we demonstrate the effectiveness of each individual module in our proposed method, establishing state-of-the-art 3D MOT performance on the KITTI and nuScenes datasets.

{\small
\bibliographystyle{ieee_fullname}
\bibliography{main}
}

\end{document}